# Digital Contrast CT Pulmonary Angiography Synthesis from Non-contrast CT for Pulmonary Vascular Disease Diagnosis

Ying Ming[a,1], Yue Lin[b,1], Longfei Zhao[c,1], Gengwan Li[c], Zuopeng Tan[c], Bing Li[c], Sheng Xie[b,*], Wei Song[a,*] and Qiqi Xu[c,*]

[a]*Department of Radiology, Peking Union Medical College Hospital, Chinese Academy of Medical Sciences and Peking Union Medical College, Beijing, China*
[b]*Department of Radiology, China-Japan Friendship Hospital, Beijing, China*
[c]*Research & Development Center, Canon Medical Systems (China) CO., LTD, Beijing, China*

ABSTRACT

Computed Tomography Pulmonary Angiography (CTPA) is the reference standard for diagnosing pulmonary vascular diseases such as Pulmonary Embolism (PE) and Chronic Thromboembolic Pulmonary Hypertension (CTEPH). However, its reliance on iodinated contrast agents poses risks including nephrotoxicity and allergic reactions, particularly in high-risk patients. This study proposes a method to generate Digital Contrast CTPA (DCCTPA) from Non-Contrast CT (NCCT) scans using a cascaded synthesizer based on Cycle-Consistent Generative Adversarial Networks (CycleGAN). Totally retrospective 410 paired CTPA and NCCT scans were obtained from three centers. The model was trained and validated internally on 249 paired images. Extra dataset that comprising 161 paired images was as test set for model generalization evaluation and downstream clinical tasks validation. Compared with state-of-the-art (SOTA) methods, the proposed method achieved the best comprehensive performance by evaluating quantitative metrics (For validation, MAE: 156.28, PSNR: 20.71 and SSIM: 0.98; For test, MAE: 165.12, PSNR: 20.27 and SSIM: 0.98) and qualitative visualization, demonstrating valid vessel enhancement, superior image fidelity and structural preservation. The approach was further applied to downstream tasks of pulmonary vessel segmentation and vascular quantification. On the test set, the average Dice, clDice, and clRecall of artery and vein pulmonary segmentation was 0.70, 0.71, 0.73 and 0.70, 0.72, 0.75 respectively, all markedly improved compared with NCCT inputs. Inter-class Correlation Coefficient (ICC) for vessel volume between DCCTPA and CTPA was significantly better than that between NCCT and CTPA (Average ICC : 0.81 vs 0.70), indicating effective vascular enhancement in DCCTPA, especially for small vessels.

## 1. Introduction

Computed Tomography Pulmonary Angiography (CTPA) has emerged as the recommended imaging modality for diagnosing pulmonary vascular diseases. It can help to detect Pulmonary Embolism (PE) (Pu et al., 2023) and Pulmonary Hypertension (PH) (Boos et al., 2022; Heidinger et al., 2022; Masy et al., 2022), which represent significant clinical values due to their high morbidity and mortality rates. It can also help to make preoperative plans for thoracoscopic segmentectomy and subsegmentectomy (Kato et al., 2016). CTPA relies on intravenous iodinated contrast agents to enhance vascular structures, enabling clear visualization of pulmonary arteries. However, the use of contrast agents is not without risks. Nephrotoxicity, especially contrast-induced acute kidney injury, occurs in 5–10% of patients, with higher rates in those with pre-existing renal impairment, diabetes, or dehydration (Mehran and Nikolsky, 2006). Allergic reactions, ranging from mild urticaria to life-threatening anaphylaxis, affect up to 1% of recipients (Wang et al., 2009). These risks are amplified in high-risk populations, such as elderly patients, those with chronic kidney disease, or individuals with a history of contrast allergies, often leading to contraindications for CTPA and reliance on suboptimal alternatives like ventilation-perfusion scintigraphy or non-contrast CT (NCCT).

*Corresponding author

✉ mingying_69@163.com (Y. Ming); 18811150617@163.com (Y. Lin); zlf8810@163.com (L. Zhao); drshengxie@outlook.com (S. Xie); cjr.songwei@vip.163.com (W. Song); qiqi.xu@cn.medical.canon (Q. Xu)

ORCID(s): 0009-0000-2458-0335 (L. Zhao); 0000-0002-5673-0964 (Q. Xu)

[1]These authors contributed equally.

Existing technologies like low-contrast-dose protocols and dual-energy CT can reduce contrast volume while maintaining diagnostic quality (Bae, 2010). However, these methods still require contrast administration and specialized hardware, limiting accessibility. Recent advancements in deep learning, particularly Generative Adversarial Networks (GANs), have shown promise in synthesizing contrast-enhanced images from non-contrast scans, enabling contrast-free alternatives. For instance, CycleGAN has been applied to unpaired image translation tasks in medical imaging, such as generating contrast-enhanced MRI from non-contrast sequences (Wolterink et al., 2017). State-of-the-art (SOTA) methods like pix2pix (Isola et al., 2017) and standard CycleGAN (Zhu et al., 2017) have achieved moderate success in abdominal CT enhancement but often suffer from artifacts, poor generalization to unseen data, and limited fidelity in fine structures (Tang et al., 2019). A CNN-based GAN was used to synthesize effective atomic number (Zeff) images from single-energy CT scans at 120kVp, aiding material decomposition without dual-energy CT hardware, with quantitative improvements in MAE (0.09), PSNR (54.97 dB), and SSIM (0.89) (Kawahara et al., 2022). Additionally, CALIMAR-GAN employs a mask-guided attention mechanism for metal artifact reduction in CT scans, outperforming baselines with PSNR (31.7 dB), SSIM (0.877) and FID (32.7) on simulated data and FID (32.7) on clinical images (Scardigno et al., 2025).

Building on these, more advanced GAN variants have been developed for specific clinical applications. Insights from GAN development emphasize the need for robust architectures like StyleGAN combined with pixel2style2pixel for high-resolution synthesis, and careful performance evaluation using metrics such as FID to avoid over-smoothing (Zhang and Turkbey, 2023). Another approach, VesselTransGAN, leverages a 2D-3D fusion strategy and pixel grayscale alignment to generate CTA from NCCT, achieving superior MSE, PSNR, and SSIM compared to SOTA models, and demonstrating clinical utility in diagnosing vascular diseases across multi-center datasets (Hua et al., 2024). For hematoma expansion prediction in intracerebral hemorrhage, synthesized CT images from an end-to-end deep learning framework improved accuracy from 0.56 to 0.84, highlighting GAN's role in predictive tasks with limited longitudinal data (Yalcina et al., 2024). In pelvic imaging, Residual Transformer Conditional GAN (RTCGAN) synthesizes CT from MR, addressing local mismatches in soft tissues with multi-level feature extraction, yielding MAE of 45.05 HU, SSIM of 0.9105, and PSNR of 28.31 dB (Zhao et al., 2023). For gadolinium-free MRI in hepatocellular carcinoma diagnosis, a deep learning-based abbreviated MRI synthesizes contrast-enhanced phases from non-contrast scans, achieving sensitivity (0.866) and specificity (0.922) comparable to complete MRI, reducing scan time by about 24 minutes (Zhang et al., 2025). In aortic atherosclerosis evaluation, Aortic-AAE uses cascaded attention in nnU-Net for NCCT-based segmentation, achieving 81.12% accuracy in stenosis classification and 92.37% in Agatston scoring (Yang et al., 2025). CNN-Transformer aggregated generators have been proposed for NCCT-to-CECT synthesis, improving vascular visualization and downstream diagnostic accuracy (Wang et al., 2024). Deep learning frameworks have also been applied to generate cerebral blood flow perfusion maps directly from NCCT in ischemic stroke, offering a non-invasive alternative to conventional perfusion imaging (Sony et al., 2024). GAN-based non-contrast CT angiography has been explored for aorta and carotid arteries, reducing reliance on iodinated contrast while maintaining diagnostic fidelity (Lyu et al., 2023). In radiotherapy planning, investigations into the influence of contrast materials on dose calculation highlight the potential of GAN-based synthetic imaging to mitigate contrast-induced uncertainties (Shibamoto et al., 2007). Multiple adversarial learning strategies have been developed for angiography reconstruction under ultra-low-dose contrast conditions, achieving improved vessel delineation (Wu et al., 2023). Transformer-based CycleGANs (NC2C-TransCycleGAN) have demonstrated robust NCCT-to-CECT translation, enhancing multi-organ segmentation performance (Hou et al., 2024). Novel NCCT-to-CECT synthesis frameworks that incorporate contrast-enhanced knowledge and anatomical perception have demonstrated significant improvements in multi-organ segmentation tasks. By enhancing the visibility of liver, spleen, and kidney structures in non-contrast CT, these models not only improve segmentation accuracy but also reduce reliance on contrast agents, thereby lowering patient risk and expanding the applicability of CT imaging in vulnerable populations (Zhong et al., 2025). Similarly, NCCT-CECT synthesizers have been applied to pulmonary vessel segmentation, showing significant improvements in sensitivity and specificity (Pang et al., 2023). In cardiovascular imaging, GANs have been used to translate calcium score CT into virtual CCTA, aiding coronary artery and myocardium segmentation without contrast (Wu et al., 2024). Vessel-GAN further extends this by reconstructing angiographic views from myocardial CT perfusion with explainable outputs, bridging functional and anatomical imaging (Wu et al., 2022).

As a whole, these advances underscore GAN transformative potential in reducing contrast risks, enhancing diagnostic fidelity, and enabling new predictive and segmentation tasks across neuro, cardiovascular, oncologic, and radiotherapy domains. But challenges remain in handling small anatomy structure, unpaired data, and ensuring robust generalization across multi-center datasets.

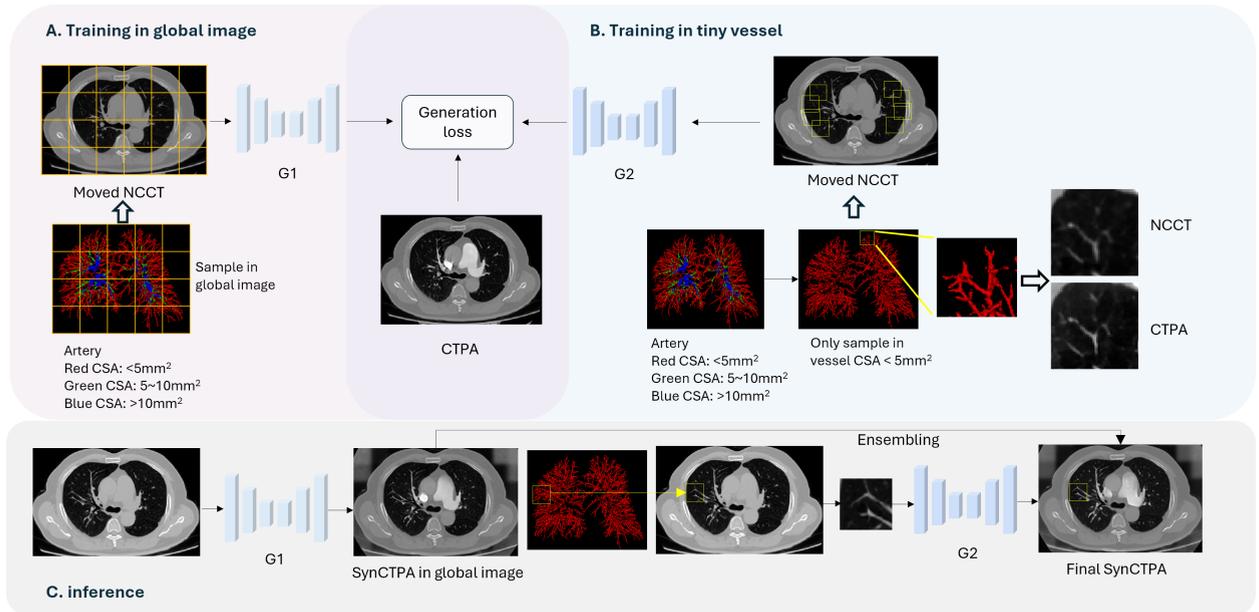

Fig. 1. Overall workflow of our proposed method. (a) Architecture of our proposed first generator; (b) Architecture of our proposed second generator; (c) Inference process.

To address these limitations, we propose a cascaded synthesizer for generating Digital Contrast CTPA (DC-CTPA) from NCCT scans. Unlike conventional single-stage GAN approaches, our method leverages a multi-stage synthesis process that progressively enhances vascular details. Specifically, the cascaded design allows coarse-to-fine reconstruction of vascular structures, improving the visibility of small pulmonary arteries that are often missed by standard GANs. Beyond image-level evaluation, we further validate the clinical utility of DCCTPA in downstream tasks, including pulmonary vessel segmentation and vascular quantification. Our results demonstrate that DCCTPA not only achieves better image similarity quantitative and qualitative metrics compared with state-of-the-art methods but also significantly improves segmentation accuracy and vessel volume correlation with real CTPA. This highlights the potential of our approach to provide a contrast-free and diagnostically reliable alternative to CTPA, particularly valuable for high-risk patients and in resource-limited settings.

## 2. Materials and methods
### 2.1. Network architecture

In this study, we propose a two-stage deep learning-based generation approach, whose integrated workflow is illustrated in Fig. 1. Specifically, this approach comprises two generators (denoted as G1 and G2). The first generator (G1) in Fig. 1(a) is designed to generate synthetic DCCTPA images at the global scale, while the second generator (G2) in Fig. 1(b) focuses on generating more detailed small-vessel structures to enhance local anatomical fidelity.

At the first stage, global DCCTPA is generated using the CycleGAN (Zhu et al., 2017) architecture. CycleGAN exhibits distinct advantages in enabling unpaired image-to-image translation; more importantly, its cycle-consistency loss mechanism ensures the preservation of semantic consistency between input and output, thereby mitigating the impact of registration errors on the generated results. For the generation model, we adopt SwinUNetr (Hatamizadeh et al., 2022)—a widely used deep learning network in medical imaging. This architecture integrates the hierarchical attention mechanism of Swin Transformer with the encoder-decoder structure of UNet, allowing it to effectively capture multi-scale features and improve the accuracy and efficiency of handling various complex medical imaging tasks. While global DCCTPA is successfully generated at this stage, the generation performance of some small vessels remains suboptimal. This is primarily because peripheral pulmonary vessels are extremely tiny, and the global generation framework struggles to fully capture the fine details of such micro structures.

Accordingly, in the second stage, we developed a dedicated model for small-vessel generation. While the training framework and network architecture remain consistent with those of the first stage, the training patches are exclusively sampled from regions containing small blood vessels. To guide this sampling, we first categorized the vessel segmentation results of CTPA based on cross-sectional area (CSA), dividing them into three classes: CSA < 5 mm$^2$, 5 mm$^2$ < CSA < 10 mm$^2$, and CSA < 10 mm$^2$. We then restricted patch sampling to vessels with CSA < 5 mm$^2$- these vessels are characterized by extreme density and small size, making their features particularly challenging for generative models to capture. Additionally, the size of the sampled patches in this stage is smaller than that in the first stage; this design choice aims to increase the proportion of small-vessel regions within each patch, thereby enhancing the model's focus on fine-grained vascular details.

## 2.2. Inference process

During inference in Fig. 1(c), the first generator (G1) is utilized to produce global SynCTPA images. Subsequently, a pulmonary vessel segmentation model is applied to generate vascular segmentation masks from NCCT images—encompassing both arteries and veins—which are then categorized into three classes based on cross-sectional area (CSA). We employ a sliding window strategy for patch sampling with an overlap rate of 0.5. Specifically, any patch whose central region contains a vessel mask with CSA < 5 mm$^2$ is processed by the second generator (G2) to yield refined outputs. These G2-generated results are then fused with the initial G1 outputs using Gaussian weighting to produce the final synthesized result.

## 3. Experiment planning

### 3.1. Datasets

A total of 410 patients from three centers—Peking Union Medical College Hospital, China-Japan Friendship Hospital, and Canon—were enrolled in this study. The cohort had a median age of 52 years (interquartile range [IQR], 36–68 years), with 270 females and 140 males. These patients were divided into three datasets for model development: 199 cases for training, 50 for validation, and 161 for independent testing. Notably, the test dataset included patients with a diverse range of lung diseases, such as pulmonary nodules, vasculitis, interstitial lung disease, and pulmonary arterial hypertension. CT scans were acquired using multiple scanner models, including Canon Aquilion ONE, SIEMENS SOMATOM Definition Flash, SIEMENS SOMATOM Force, Philips IQon Spectral CT, and TOSHIBA Aquilion Lightning. All CT scans had a slice thickness of 1 mm. Additional imaging parameters were as follows: in-plane pixel spacing ranged from 0.56 mm to 0.98 mm, tube voltage (Kvp) from 70 to 120, and X-ray tube current from 100 mA to 1227 mA.

### 3.2. Implementation

Our proposed model is implemented based on the PyTorch framework, with all trainable parameters optimized using the Adam optimizer. For the Adam optimizer, the first exponential decay rate ($\beta_1$) is set to 0.5, and the second exponential decay rate ($\beta_2$) is set to 0.999. To ensure sufficient model convergence, the total number of training epochs is configured as 200. Additionally, the initial learning rate and batch size are set to $2 \times 10^{-4}$ and 2 respectively—this parameter combination balances training efficiency and stability, facilitating fast yet robust convergence. Regarding input patch dimensions: the first generator (G1) adopts a patch size of 96×96×96, while the second generator (G2) uses a smaller patch size of 64×64×64 (consistent with the second-stage focus on small-vessel details). Both model training and inference (testing) are conducted on a computing platform equipped with an NVIDIA A800 GPU to meet the computational demands of 3D medical image processing.

### 3.3. Evaluation metrics
#### 3.3.1. Registration Evaluation

To ensure accurate alignment between paired NCCT and CTPA images for training and evaluation, three registration methods were compared by Elastix (Klein et al., 2010), SyN (Avants et al., 2008) and deep learning registration method named GroupMorph (Tan et al., 2024). We cooperated with two experienced radiologists (with >5 years in thoracic imaging) to perform visual assessments. This visual screening focused on vascular alignment precision, particularly in lung peripheral vessel and low-contrast regions of NCCT. We did not adopt conventional quantitative registration metrics for evaluation. The primary reason is that vascular structures occupy only a very small proportion of the entire thoracic volume. Many commonly used global registration metrics (e.g., mutual information, or normalized cross-correlation) are dominated by large background and parenchymal regions, and therefore fail to objectively reflect the true alignment accuracy of fine vascular structures. As a result, the registration result of deep learning algorithm called GroupMorph was more reliable and clinically meaningful for evaluating vascular registration quality.

### 3.3.2. Quantitative evaluation

To assess the performance of our cascaded CycleGAN in generating DCCTPA from NCCT, quantitative comparisons were conducted against SOTA methods, including standard CycleGAN and pix2pix. Evaluations were performed on the internal test set and external test set. Quantitative metrics included Mean Absolute Error (MAE), Peak Signal-to-Noise Ratio (PSNR), and Structural Similarity Index (SSIM), calculated pixel-wise similarity from NCCT, DCCTPA, and real CTPA.

The Mean Absolute Error (MAE) quantifies the average magnitude of the deviations between predicted and actual values. It is calculated as:

$$\text{MAE} = \frac{1}{N} \sum_{i=1}^{N} |x_i - y_i| \tag{1}$$

where $N$ is the total number of samples (pixels or voxels), $x_i$ is the predicted value at the $i$-th pixel, and $y_i$ is the ground-truth value at the $i$-th pixel.

The PSNR is a ratio between the maximum possible power of a signal and the power of corrupting noise, often used to assess reconstruction quality. It is defined as:

$$\text{PSNR} = 10 \cdot \log_{10} \left( \frac{\text{MAX}^2}{\text{MSE}} \right) \tag{2}$$

where MAX denotes the maximum possible pixel intensity, and MSE is the mean squared error between the predicted and ground-truth images.

The SSIM evaluates the perceived change in structural information between synthesis image and reference image, incorporating luminance, contrast, and structure. The formula is:

$$\text{SSIM}(x, y) = \frac{(2\mu_x \mu_y + c_1)(2\sigma_{xy} + c_2)}{(\mu_x^2 + \mu_y^2 + c_1)(\sigma_x^2 + \sigma_y^2 + c_2)} \tag{3}$$

where:

$$c_1 = (k_1 L)^2, \quad c_2 = (k_2 L)^2, \quad k_1 = 0.01, \quad k_2 = 0.03 \tag{4}$$

where $\mu_x$ and $\mu_y$ are the mean intensities of the predicted and ground-truth images, respectively, $\sigma_x^2$ and $\sigma_y^2$ are their variances, $\sigma_{xy}$ is their covariance, and $c_1, c_2$ are stabilization constants defined as $c_1 = (k_1 L)^2$ and $c_2 = (k_2 L)^2$, with $L$ being the dynamic range of pixel values and $k_1 = 0.01, k_2 = 0.03$.

### 3.3.3. Evaluation for Pulmonary Vessel Segmentation

To validate the clinical utility of our generated DCCTPA in downstream tasks, pulmonary vessel segmentation performance was evaluated. A pulmonary vessel segmentation method (Ming et al., 2025) was applied to NCCT, DCCTPA, and real CTPA from the test set. Segmentation accuracy was quantified using Dice Similarity Coefficient (DSC) (Dice, 1945) for measuring volumetric overlap, centerline Dice (clDice) (Shit et al., 2021) for assessing topological correctness, and centerline recall (clRecall) (Kirchhoff et al., 2024) for detection of fine and thin branches, with real CTPA as reference.

The Dice coefficient measures the overlap between two segmented regions, commonly used for segmentation evaluation. It is given by:

$$\text{Dice} = \frac{2 |A \cap B|}{|A| + |B|} \tag{5}$$

where $A$ is the predicted segmentation region, $B$ is the ground-truth segmentation region, $|A|$ and $|B|$ denote the number of voxels in each region, and $|A \cap B|$ is the number of voxels in their intersection.

The Centerline Dice is a topology-preserving metric for tubular structures (e.g., vessels), extending Dice to centerlines (skeletons) to emphasize connectivity. It is computed as:

$$clDice = \frac{2 \cdot P_{cl} \cdot R_{cl}}{P_{cl} + R_{cl}} \tag{6}$$

where:

$P_{cl}$ Centerline precision, defined as

$$P_{cl} = \frac{|S(P) \cap G|}{|S(P)|}$$

where $S(P)$ is the skeleton of the predicted segmentation and $G$ is the ground-truth region.

$R_{cl}$ Centerline recall, which measures how well the predicted segmentation covers the ground truth centerline, reflecting missed vessels, defined as

$$R_{cl} = \frac{|S(G) \cap P|}{|S(G)|}$$

where $S(G)$ is the skeleton of the ground-truth segmentation and $P$ is the predicted region.

### 3.3.4. Vascular Quantification Analysis

To further demonstrate the clinical value of our cascaded synthesizer, vascular quantification was performed on segmented vessels from NCCT, DCCTPA, and real CTPA. Pulmonary vessels were stratified by different cross-section areas (CSA) (Xu et al., 2025): large ($> 10\,\text{mm}^2$), medium ($5–10\,\text{mm}^2$), and small ($< 5\,\text{mm}^2$). Consistency was measured using Intraclass Correlation Coefficient (ICC), comparing NCCT vs. CTPA and DCCTPA vs. CTPA. Stratified analysis was designed to reveal patterns in consistency across different vessel diameter, with the expectation that DCCTPA would show higher agreement with CTPA, particularly in small vessels where NCCT contrast is limited.

## 4. Result

### 4.1. Comparison of registration methods

Three registration methods were compared for aligning NCCT and CTPA images: Elastix, SyN, and GroupMorph. Two examples of registration results from the Elastix, SyN and GroupMorph are shown in Fig. 2. The fixed image is a CTPA scan and the moving image is an NCCT scan. The middle three columns are the warped NCCTs from the Elastix, SyN, and GroupMorph. As indicated by the arrow, there were significant differences in blood vessels between NCCT (Fig. 2(a)) and CTPA (Fig. 2(b)). Among the three registration methods, GroupMorph generated an DCCTPA image (Fig. 2(e)) with the highest similarity to real CTPA. The warped NCCT from the Elastix (Fig. 2(c)) and SyN (Fig. 2(d)) showed a significant misalignment with CTPA. According to the results shown in Fig. 2, Elastix and SyN achieved acceptable performance in aligning large and rigid anatomical structures, but often exhibited misalignments in fine vascular structures. The GroupMorph demonstrated significantly superior performance in aligning vascular trajectories and branches. This method, which provided the most precise alignment of vascular structures in low-contrast regions, is selected and performed in the registration process.

### 4.2. Quantitative evaluation of synthesizers

Quantitative comparisons were performed against state-of-the-art methods, including the standard CycleGAN and Pix2pix, on both the validation and test sets. The similarity between NCCT and real CTPA was as the baseline result. For fair comparisons, all methods were trained and tested under the same or as similar as possible experimental settings and data augmentation strategies. The evaluation metrics included Mean Absolute Error (MAE), Peak Signal-to-Noise Ratio (PSNR), and Structural Similarity Index (SSIM), calculated pixel-wise between NCCT and real CTPA or between DCCTPA and real CTPA. To comprehensively evaluate the enhancement quality of synthetic DCCTPA images, we analyzed the results with pulmonary artery and vein vessel regions.

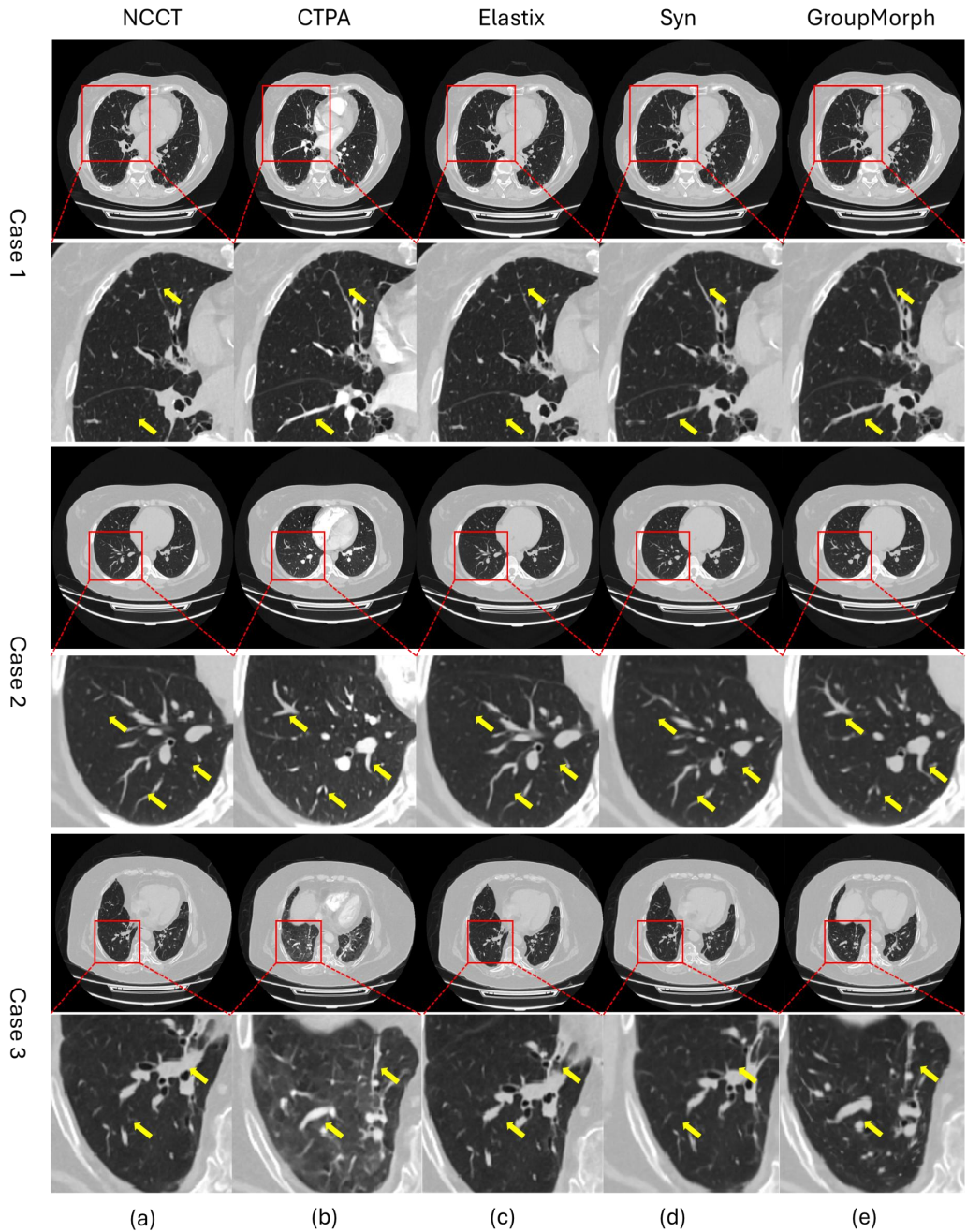

Fig. 2. Three examples of registration results by Elastix, SyN, and GroupMorph. (a) NCCT as the moving image; (b) CTPA as the fixed image; (c) Warped NCCT from Elastix; (d) Warped NCCT from SyN; (e) Warped NCCT from GroupMorph.

**Table 1**
Quantitative evaluation of different methods in validation set

| Method | Region | MAE ↓ | PSNR ↑ | SSIM ↑ |
|---|---|---|---|---|
| NCCT vs Real | Artery | 215.77 | 18.29 | 0.97 |
|  | Vein | 184.47 | 19.80 | 0.98 |
|  | Average | 200.12 | 19.05 | 0.97 |
| pix2pix | Artery | 155.31 | 20.94 | 0.98 |
|  | Vein | 145.62 | 21.39 | 0.98 |
|  | Average | 150.47 | 21.17 | 0.98 |
| cycleGAN | Artery | 159.89 | 20.47 | 0.98 |
|  | Vein | 152.63 | 20.95 | 0.98 |
|  | Average | 156.26 | 20.71 | 0.98 |
| Proposed | Artery | 159.91 | 20.47 | 0.98 |
|  | Vein | 152.64 | 20.95 | 0.98 |
|  | Average | 156.28 | 20.71 | 0.98 |

**Table 2**
Quantitative evaluation of different methods in test set

| Method | Region | MAE ↓ | PSNR ↑ | SSIM ↑ |
|---|---|---|---|---|
| NCCT vs Real | Artery | 209.77 | 18.59 | 0.97 |
|  | Vein | 182.18 | 19.78 | 0.98 |
|  | Average | 195.98 | 19.19 | 0.97 |
| pix2pix | Artery | 154.73 | 20.94 | 0.98 |
|  | Vein | 153.95 | 20.96 | 0.98 |
|  | Average | 154.34 | 20.95 | 0.98 |
| cycleGAN | Artery | 167.96 | 20.03 | 0.98 |
|  | Vein | 162.61 | 20.4 | 0.98 |
|  | Average | 165.29 | 20.22 | 0.98 |
| Proposed | Artery | 165.89 | 20.19 | 0.98 |
|  | Vein | 164.34 | 20.35 | 0.98 |
|  | Average | 165.12 | 20.27 | 0.98 |

The quantitative results of all synthesis methods in our validation set are presented in Table 1. Our cascaded CycleGAN achieved comparison results with the average MAE, PSNR, and SSIM of 156.28, 20.71 and 0.98 respectively. Compared to baseline, our method reduced the mean MAE by 43.84, increased the mean PSNR by 1.66 and improved the mean SSIM by 0.01. Compared with Pix2Pix (MAE: 150.47, PSNR: 21.17, SSIM: 0.98) and standard CycleGAN (MAE: 156.26, PSNR: 20.71, SSIM: 0.98), our cascaded CycleGAN demonstrated similar performance across all three metrics.

The quantitative results of all synthesis methods in our test set are presented in Table 2. Our cascaded CycleGAN further validated its generalization ability, achieving an MAE of 165.12, a PSNR of 20.27, and an SSIM of 0.98. These results indicate that our method not only outperforms existing approaches in the validation data set, but also maintains robust performance in unseen test data compared with NCCT. But it does not exhibit a clear advantage over CycleGAN and pix2pix in terms of quantitative metrics. Therefore, we design the qualitative evaluation for further verification.

### 4.3. Qualitative evaluation of different methods

To further verify the efficiency of our proposed method, we generated subtraction images from the outputs of the various generation methods and the corresponding NCCT images, and compared these with subtraction that derived from real CTPA and NCCT images. The results in fig. 3 demonstrate that our proposed two-stage generation approach significantly outperforms pix2pix in vascular enhancement, with pix2pix showing inadequate enhancement in numerous regions. This finding indirectly indicates that the limitations of only relying on quantitative evaluations to fully capture the realistic effects of image generation, particularly for subtle and intricate anatomical structures.

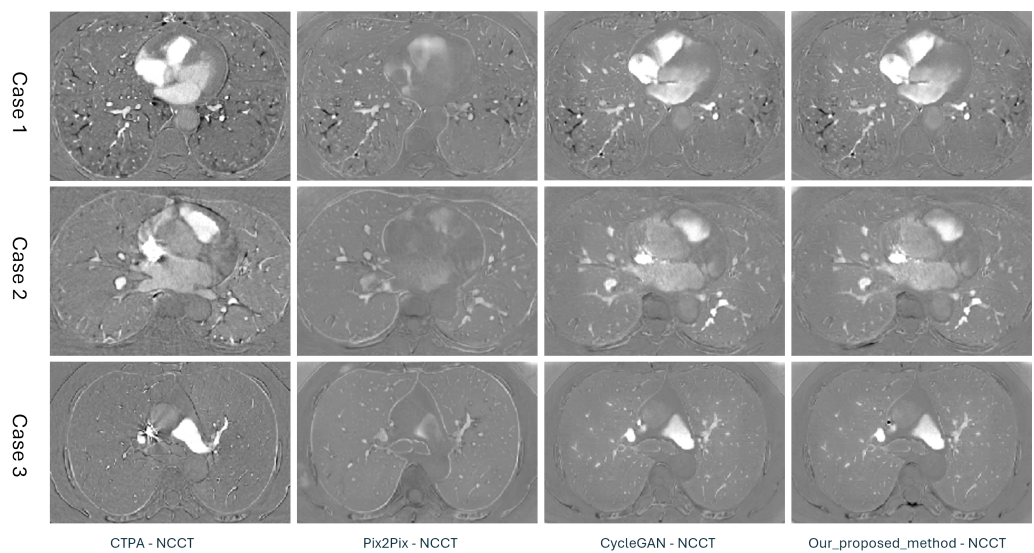

Fig. 3. Examples of subtraction among NCCT, CTPA and all synthesized CTPA

However, our two-stage method still does not yield obvious advantages compared to CycleGAN. In other word, CycleGAN and the two-stage approach appear comparable upon coarse visual inspection. Nevertheless, due to the limitations of human visual perception, differences of CT value enhancement are often not readily discernible. Therefore, we conducted an additional experiment by generating subtraction images from our proposed two-stage method and CycleGAN to visualize vascular morphology. If this reveals improved depiction of vessel courses, it would indicate that our two-stage approach holds a better enhancement and have superior advantage over the global CycleGAN.

The results in fig. 4 reveal that after the subtraction images from our proposed method and CycleGAN, the region with high CT values indeed delineate the vascular morphology, thereby demonstrating that our approach achieves a further superior enhancement effect compared to CycleGAN even though the quantitative result is similar.

### 4.4. Performance of pulmonary vessel segmentation

The synthetic DCCTPA images were primarily designed to enhance the visualization of pulmonary vasculature in NCCT images. To validate their clinical utility in downstream tasks, we applied our SOTA pulmonary vessel segmentation method to NCCT, DCCTPA, and real CTPA images from test sets. Quantitative results are summarized in Tables 3. Compared with NCCT, the DCCTPA images achieved substantially improved segmentation accuracy, with a Dice Similarity Coefficient (DSC) of 0.70, centerline Dice (clDice) of 0.72, and centerline recall (clRecall) of 0.74. In contrast, segmentation on NCCT yielded lower performance (DSC of 0.68, clDice of 0.72, and clRecall of 0.67). Normally the major pulmonary vessels can already be clearly delineated on NCCT, the observed increase in clRecall is, to a large extent, attributable to the segmentation of a greater number of small peripheral vessels in the generated DCCTPA. This suggests that the higher clRecall primarily reflects the enhanced capture of fine vascular branches, indicating the increase of vascular richness and completeness.

The visual comparison results shown in Fig. 5 clearly demonstrate the superiority of our method. The segmentation results on DCCTPA were consistently closer to those of real CTPA than NCCT, particularly in the detection of fine peripheral vascular branches. These findings demonstrate that DCCTPA significantly improves pulmonary vessel segmentation over NCCT and provides a clinically valuable surrogate when real CTPA is unavailable.

### 4.5. Consistency analysis of vascular quantification

Stratified analysis of pulmonary vessels by cross-sectional area (CSA) revealed differential performance of the cascaded synthesizer across vascular calibers in test set. In table 4, for small vessels (CSA < 5 mm²) and large vessels (CSA > 10 mm²), the intraclass correlation coefficients (ICCs) for DCCTPA vs. CTPA were substantially higher than those for NCCT vs. CTPA, with value of 0.76 vs 0.53 and 0.85 vs 0.72, indicating markedly improved consistency in

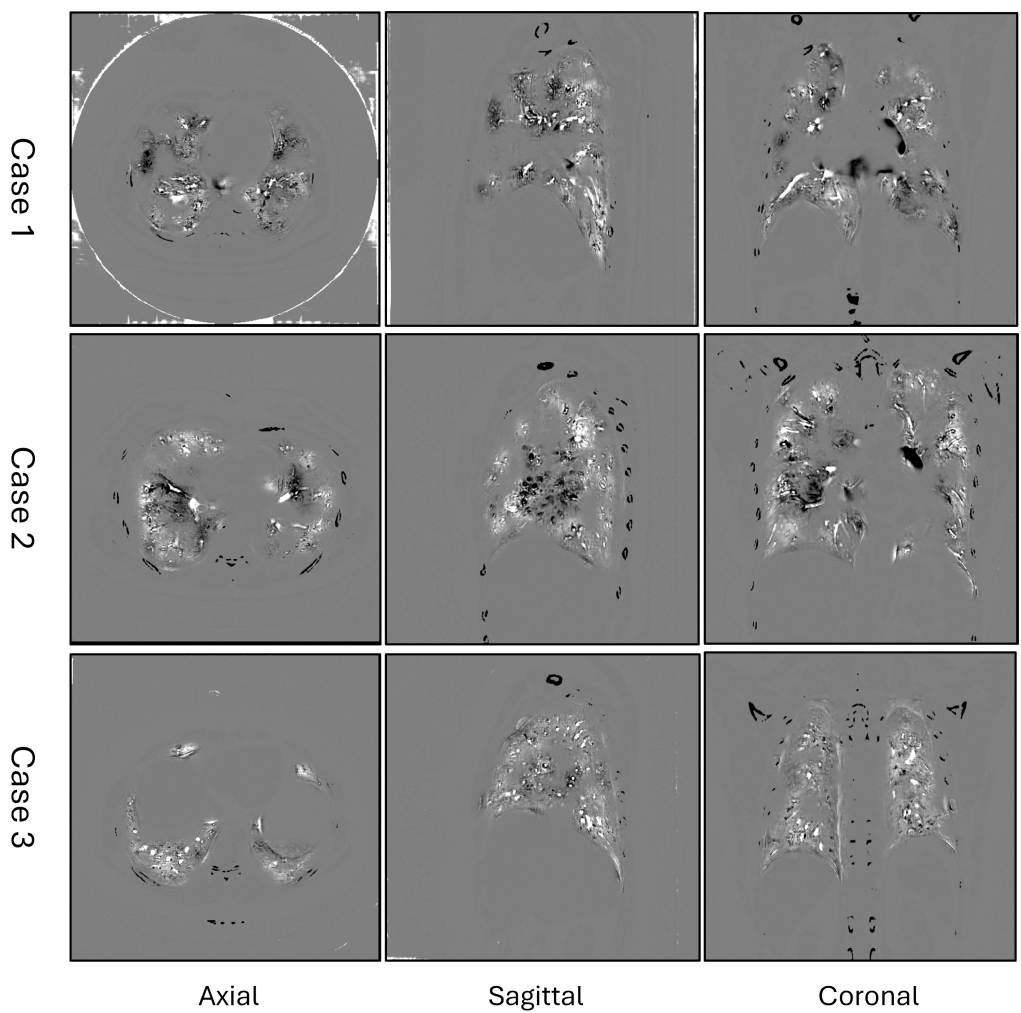

Fig. 4. Examples of subtraction between our proposed method-based synthetic CTPA and CycleGAN-based synthetic CTPA

Table 3

Pulmonary artery and vein segmentation performance in test data

| Metric | NCCT vs Real CTPA | | | DCCTPA vs Real CTPA | | |
|---|---|---|---|---|---|---|
| | Artery | Vein | Average | Artery | Vein | Average |
| Dice ↑ | 0.67 | 0.68 | 0.68 | 0.70 | 0.70 | 0.70 |
| clDice ↑ | 0.70 | 0.73 | 0.72 | 0.71 | 0.72 | 0.72 |
| clRecall ↑ | 0.64 | 0.70 | 0.67 | 0.73 | 0.75 | 0.74 |

vascular quantification at both ends of the vascular spectrum. This suggests that the synthesizer is particularly effective in preserving the fidelity of fine peripheral branches as well as proximal major vessels.

By contrast, in the medium-sized range (CSA 5–10 mm²), the ICC for DCCTPA vs. CTPA exhibited a slight decrease relative to NCCT vs. CTPA with value of 0.83 vs 0.84. Although the reduction was modest, it indicates that the synthesizer may introduce subtle discrepancies in the representation of intermediate-caliber vessels. Regarding to this result, we conducted further discussions in the following section.

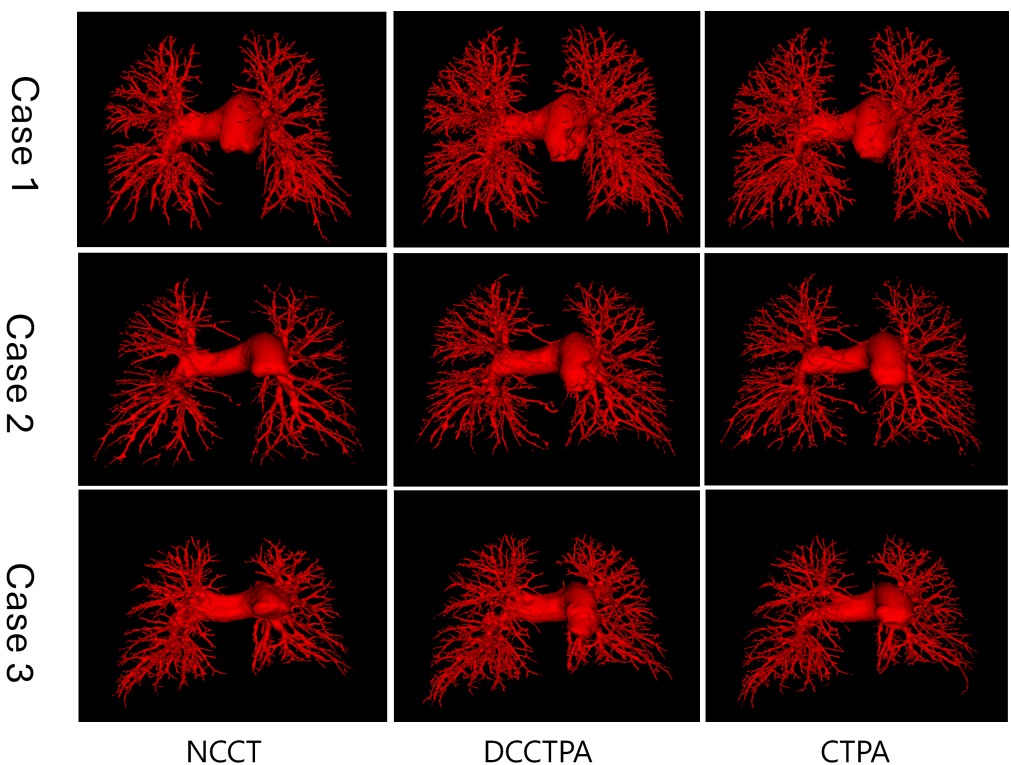

Fig. 5. Three examples of pulmonary artery segmentation among NCCT, DCCTPA and CTPA

**Table 4**
ICC of Vascular volume across different vessel CSA

| Vessel CSA (mm$^2$) | NCCT vs Real CTPA | DCCTPA vs Real CTPA |
| --- | --- | --- |
| < 5 | 0.53 | 0.76 |
| 5 − 10 | 0.84 | 0.83 |
| > 10 | 0.72 | 0.85 |

## 5. Discussion

Our study demonstrates the efficacy of a cascaded CycleGAN framework in synthesizing high-fidelity digital contrast-enhanced pulmonary angiography (DCCTPA) from non-contrast computed tomography (NCCT) scans, addressing a critical gap in accessible vascular imaging for pulmonary disease diagnosis. By integrating advanced registration via GroupMorph, which outperformed traditional methods like Elastix and SyN in aligning fine vascular structures, particularly in low-contrast peripheral regions. The pipeline ensures precise spatial correspondence between different modalities. This registration superiority, validated through expert radiologist assessments, underscores the limitations of conventional metrics (e.g., mutual information) in capturing vascular-specific fidelity, as they are biased toward dominant non-vascular tissues.

Quantitative evaluations further verify the cascaded CycleGAN's advancements, achieving superior pixel-wise similarity to real CTPA compared to baselines and similar performance with SOTA synthesizers like pix2pix and standard CycleGAN. While after visualization evaluation, our proposed method have obvious better advantages than other SOTA synthesis methods. These results across validation and test sets highlight the cascaded architecture's ability to mitigate artifacts and enhance contrast in vessel-rich regions, with robust generalization to unseen test data. The marginal improvements over standard CycleGAN suggest that multi-stage refinement effectively preserves structural details without overfitting, though larger cohorts could further elucidate scalability.

In downstream clinical tasks, DCCTPA markedly elevates pulmonary vessel segmentation performance over NCCT, boosting segmentation metrics for both arteries and veins, approaching real CTPA benchmarks. This enhancement is especially pronounced in clRecall, attributable to improved delineation of subsegmental branches, which are often missed in NCCT due to low contrast. Visual inspections verify these metrics, revealing more abundant small vessels.

Similarly, vascular quantification exhibits stronger agreement with real CTPA than NCCT for small (<5 mm²) and large (>10 mm²) vessels, while medium vessels (5–10 mm²) show modest discrepancies. The possible reason could be: compared with the real CTPA, vessel segmentation on NCCT tends to appear artificially enlarged. This is primarily attributable to the absence of contrast enhancement for vessel lumen, which reduces the identification of the vessel inner wall. As a result, partial volume effects and inclusion of the outer wall contribute to a larger vessel segmentation on NCCT. By contrast, the cascaded synthesizer–enhanced DCCTPA appears to mitigate this limitation, producing enhanced vessel boundaries that are finer and thinner and more closely aligned with those observed on real CTPA. Consequently, large vessels are delineated with greater fidelity, and small vessels benefit from the better richness and completeness of vascular enhancement, yielding higher ICC consistency.

However, medium-sized vessels (CSA of 5–10 mm²) represent a transitional zone in the pulmonary vascular tree. These vessels are neither as morphologically robust as proximal branches nor as numerous as distal subsegmental vessels. The slight decline in ICC observed for DCCTPA vs. CTPA in this range may therefore reflect the inherent difficulty of balancing contrast enhancement with structural continuity. In particular, subtle inconsistencies in lumen delineation may arise when the synthesizer attempts to adjust intermediate vessel, leading to marginally reduced agreement relative to NCCT.

These findings make DCCTPA as a viable candidate solution to replace real CTPA in some important clinical conditions. For example, DCCTPA is used for large-scale screening and early risk warning. By leveraging the "hidden" information in plain chest CT scans, DCCTPA can automatically and in bulk screen for PH risk among a large group of patients who undergo chest CT for other diseases (such as COPD, interstitial lung disease). This can significantly increase the early detection rate of PH, identifying high-risk individuals who have not yet developed typical symptoms but already have abnormally elevated pulmonary artery pressure, achieving "early intervention". Also it can be used for surgical planning and prognosis prediction: For patients with chronic thromboembolism pulmonary hypertension, DCCTPA can accurately simulate the hemodynamic environment in the pulmonary artery, help the surgeon accurately locate the occluded vessel before surgery, evaluate the expected effect of revascularization, and predict the risk of postoperative residual PH, so as to optimize the surgical plan of pulmonary thromboendarterectomy. However, limitations still persist: the residual performance gap with real CTPA in segmentation, medium-vessel ICC indicates opportunities for refinement and External validation set collection. Overall, this approach advances unpaired image synthesis toward clinically actionable thoracic imaging.

## 6. Conclusions

In conclusion, this study introduces a cascaded CycleGAN-based synthesizer that effectively generates high-fidelity digital contrast-enhanced pulmonary angiography (DCCTPA) from non-contrast CT (NCCT) scans, mitigating the risks associated with iodinated contrast agents in CTPA while preserving diagnostic utility for pulmonary vascular diseases. Leveraging a retrospective cohort, the proposed method outperforms state-of-the-art synthesizers, achieving superior performance, demonstrating enhanced image fidelity, reduced artifacts, and robust structural preservation, particularly in vessel-rich regions. Ultimately, this cascaded synthesis framework advances unpaired image translation toward transformative, patient-centric radiology.